\documentclass[10pt,twocolumn,letterpaper]{article}
\usepackage{cvpr}
\usepackage{times}
\usepackage{epsfig}
\usepackage{graphicx}
\usepackage{amsmath}
\usepackage{amssymb}
\usepackage{multirow}
\usepackage[pagebackref=true,breaklinks=true,letterpaper=true,colorlinks,bookmarks=false]{hyperref}

\cvprfinalcopy 

\def\cvprPaperID{3602} 

\ifcvprfinal\fi
\begin{document}

\title{xUnit: Learning a Spatial Activation Function for Efficient Image Restoration}

\author{Idan Kligvasser, Tamar Rott Shaham and Tomer Michaeli\\
Technion--Israel Institute of Technology,
Haifa, Israel\\
{\tt\small \{kligvasser@campus,stamarot@campus,tomer.m@ee\}.technion.ac.il}
}
\maketitle
\thispagestyle{empty}

\begin{abstract}

In recent years, deep neural networks (DNNs) achieved unprecedented performance in many low-level vision tasks. However, state-of-the-art results are typically achieved by very deep networks, which can reach tens of layers with tens of millions of parameters. To make DNNs implementable on platforms with limited resources, it is necessary to weaken the tradeoff between performance and efficiency. In this paper, we propose a new activation unit, which is particularly suitable for image restoration problems. In contrast to the widespread per-pixel activation units, like ReLUs and sigmoids, our unit implements a learnable nonlinear function with \emph{spatial} connections. This enables the net to capture much more complex features, thus requiring a significantly smaller number of layers in order to reach the same performance. We illustrate the effectiveness of our units through experiments with state-of-the-art nets for denoising, de-raining, and super resolution, which are already considered to be very small. With our approach, we are able to further reduce these models by nearly $50\%$ without incurring any degradation in performance.
\end{abstract} 

\section{Introduction}

Deep convolutional neural networks (CNNs) have revolutionized computer vision, achieving unprecedented performance in high-level vision tasks such as classification \cite{vgg,resnet,densenet}, segmentation \cite{unet,segnet} and face recognition \cite{parkhi2015deep,facenet}, as well as in low-level vision tasks like denoising \cite{formresnet,dncnn,mlp,remez2017deep}, deblurring \cite{noroozi2017motion}, super resolution \cite{srcnn,srgan,kim2016accurate} and dehazing \cite{ren2016single}. 
Today, the performance of CNNs is still being constantly improved, mainly by means of increasing the net's depth. Indeed, identity skip connections~\cite{he2016identity} and residual learning~\cite{resnet,dncnn}, used within ResNets~\cite{resnet} and DenseNets~\cite{densenet}, now overcome some of the difficulties associated with very deep nets, and have even allowed to cross the 1000-layer barrier~\cite{he2016identity}.

The strong link between performance and depth, has major implications on the computational resources and running times required to obtain state-of-the-art results. In particular, it implies that applications on real-time, low-power, and limited resource platforms (\eg mobile devices), cannot currently exploit the full potential of CNNs.

\begin{figure}
  \centering
  \includegraphics[width=\columnwidth]{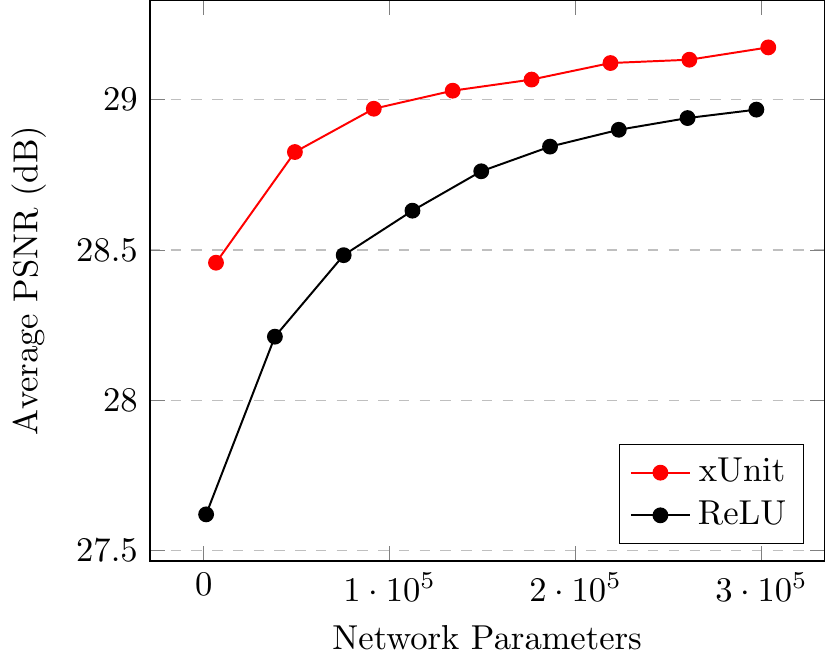}
  \caption{\textbf{xUnit activations vs.\@ ReLU activations.} xUnits are nonlinear activations with \emph{learnable spatial connections}. When used in place of the popular per-pixel activations (\eg ReLUs), they lead to significantly better performance with a smaller number of total net parameters. The graph compares the denoising performance of a conventional ConvNet (Conv+BN+ReLU layers) with our xNet (Conv+xUnit layers) at a noise level of $\sigma = 25$. As can be seen, xNet attains a much higher PSNR with the same number of parameters. Alternatively, it can achieve the same PSNR with roughly $1/3$ the number of ConvNet parameters.}\label{fig:Fig1}
\end{figure}

In this paper, we propose a different mechanism for improving CNN performance (see Fig.~\ref{fig:Fig1}). Rather than increasing depth, we focus on making the nonlinear activations more effective. 
Most popular architectures use per-pixel activation units, \eg rectified linear units (ReLUs) \cite{relu}, exponential linear units (ELUs) \cite{elu}, sigmoids \cite{orr2003neural}, etc. Here, we propose to replace those units by \emph{xUnit}, a layer with \emph{spatial} and \emph{learnable} connections. The xUnit computes a continuous-valued weight map, serving as a soft gate to its input. As we show, although it is more computationally demanding and memory consuming than a per-pixel unit, the xUnit has a dramatic effect on the net's performance. Therefore, it allows using far fewer layers to match the performance of a CNN with ReLU activations. Overall, this results in a significantly improved tradeoff between performance and efficiency, as illustrated in Fig.~\ref{fig:Fig1}.




The xUnit has a set of learnable parameters. Therefore, to conform to a total budget of parameters, xUnits must come at the expense of some of the convolutional layers in the net or some of the channels within those convolutional layers. This raises the question: What is the optimal percentage of parameters to invest in the activation units? Today, most CNN architectures are at one extreme of the spectrum, with $0\%$ of the parameters invested in the activations. Here, we show experimentally that the optimal percentage is much larger than zero. This suggests that the representational power gained by using spatial activations can be far more substantial than that offered by convolutional layers with per-pixel activations.



We illustrate the effectiveness of our approach in several image restoration tasks. Specifically, we take state-of-the-art CNN models for image denoising \cite{dncnn}, super-resolution \cite{srcnn,srgan}, and de-raining \cite{derainnet}, which are already considered to be very light-weight, and replace their activations with xUnits. We show that this allows us to further reduce the number of parameters (by discarding layers or channels) without incurring any degradation in performance. In fact, we show that for small models, we can save nearly $50\%$ of the parameters while achieving the same performance or even better. As we show, this often allows to use three orders of magnitude less training examples.


\section{Related Work}
The quest for accurate image enhancement algorithms attracted significant research efforts over the past several decades. 
Until 2012, the vast majority of algorithms relied on generative image models, usually through maximum a-posterori (MAP) estimation. Models were typically either hand-crafted or learned from training images, and included \eg priors on derivatives \cite{rudin1992nonlinear}, wavelet coefficients \cite{portilla2003image}
, filter responses \cite{foe}, image patches~\cite{bm3d,epll}, etc. In recent years, generative approaches are gradually being pushed aside by discriminative methods, mostly based on CNNs. These architectures typically directly learn a mapping from a degraded image to a restored one, and were shown to exhibit excellent performance in many restoration tasks, including
\eg denoising \cite{mlp,dncnn,formresnet}, debluring \cite{noroozi2017motion}, super-resolution \cite{dong2014learning,srcnn,srgan,kim2016accurate}, dehazing \cite{cai2016dehazenet,li2017all}, and de-raining \cite{derainnet}.


A popular strategy for improving the performance of CNN models, is by increasing their depth. Various works suggested ways to overcome some of the difficulties in training very deep nets. These opened the door to a line of algorithms using ever larger nets. Specifically, the residual net (ResNet) architecture \cite{resnet}, was demonstrated to achieve exceptional classification performance compared to a plain network. Dense convolutional networks (DenseNets) \cite{densenet} took the ``skip-connections'' idea one step further, by connecting each layer to every other layer in a feed-forward fashion. This allowed to achieve excellent performance in very deep nets.

These ideas were also adopted by the low-level vision community. In the context of denoising, Zhang \etal \cite{dncnn} were the first to train a very deep CNN for denoising, yielding state-of-the-art results. To train their net, which has $0.5$M parameters, they utilized residual learning and batch normalization~\cite{dncnn}, alleviating the vanishing gradients problem. In~\cite{formresnet}, a twice larger model was proposed, which is based on formatting the residual image to contain structured information instead of learning the difference between clean and noisy images. Similar ideas were also proposed in \cite{remez2017deep1,remez2017deep2,zhang2017learning}, leading to models with large numbers of parameters. Recently, a very deep network based on residual learning was proposed in \cite{bae2016beyond}, which contains more than 60 layers, and $17$M parameters.

In the context of super-resolution, the progress was similar. In the near past, state-of-the-art methods used only a few tens of thousands of parameter. For example, the SRCNN model \cite{srcnn} contains only three convolution layers, with only $57$K parameters. The very deep super-resolution model (VDSR) \cite{kim2016accurate} already used $20$ layers with $660$K parameters. Nowadays, much more complex models are in use. For example, the well-known SRResNet \cite{srgan} uses more than $1.5$M parameters, and the the EDSR network~\cite{lim2017enhanced} (winner of the NTIRE2017 super resolution challenge \cite{timofte2017ntire}), has $43$M parameters.



The trend of making CNNs as deep as possible, poses significant challenges in terms of running those models on platforms with low-power and limited computation and memory resources. One approach towards diminishing memory consumption and access times, is to use binarized neural networks \cite{courbariaux2016binarized}. These architectures, which were shown beneficial in classification tasks, constrain the weights and activations to be binary. Another approach is to replace the multi-channel convolutional layers by depth-wise convolutions \cite{conv2ddepthwise}. This offers a significant reduction in size and latency, while allowing reasonable classification accuracy. In \cite{shi2016real}, it was suggested to reduce network complexity and memory consumption for super resolution, by introducing a sub-pixel convolutional layer that learns upscaling filters. In~\cite{lefkimmiatis2016non}, an architecture which exploits non-local self-similarities in images, was shown to yield good results with reduced models. Finally, learning the optimal slope of leaky ReLU type activations has also shown to lead to more efficient models \cite{he2015delving}.



\section{xUnit}

\begin{figure*}
\begin{center}
\includegraphics[width=\textwidth]{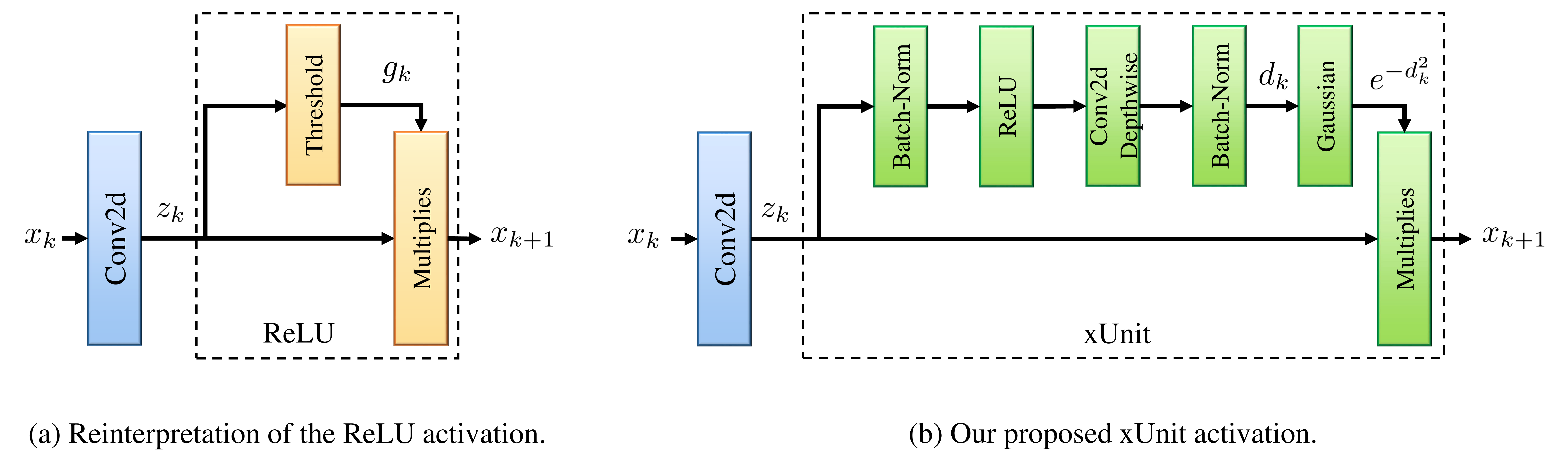}
\end{center}
\caption{\textbf{The xUnit activation layer.} (a)~The popular ReLU activation can be interpreted as performing an element-wise product between its input $z_k$ and a weight map $g_k$, which is a binarized version of $z_k$. (b)~The xUnit constructs a continuous weight map taking values in $[0,1]$, by performing a nonlinear \emph{learnable spatial function} on $z_k$.}\label{fig:scheme} 
\end{figure*}

Although a variety of CNN architectures exist, their building blocks are quite similar, mainly comprising of convolutional layers and per-pixel activation units.
Mathematically, the features $x_{k+1}$ at layer $k+1$ are commonly calculated as
\begin{align}\label{eq:CNN}
&z_k= W_{k}x_{k}+b_{k}, \nonumber\\
&x_{k+1}=f(z_k),
\end{align}
where $x_0$ is the input to the net, $W_k$ performs a convolution operation, $b_k$ is a bias term, $z_k$ is the output of the convolutional layer, and $f(\cdot)$ is some nonlinear activation function which operates element-wise on its argument. Popular activation functions include the ReLU~\cite{relu}, leaky ReLU~\cite{maas2013rectifier}, ELU~\cite{elu}, tanh and sigmoid functions.

Note that there is a clear dichotomy in \eqref{eq:CNN}: The convolutional layers are responsible for the \emph{spatial processing}, and the activation units for the \emph{nonlinearities}. One may wonder if this is the most efficient way to realize the complex functions needed in low-level vision. In particular, is there any reason not to allow spatial processing also within the activation functions?

Element-wise activations can be thought of as nonlinear gating functions. Specifically, assuming that $f(0)=0$, as is the case for all popular activations, \eqref{eq:CNN} can be written as
\begin{align}\label{eq:CNN2}
x_{k+1}=z_k\circ g_k,
\end{align}
where $\circ$ denotes the (element-wise) Hadamard product, and $g_k$ is a (multi-channel) weight map that depends on $z_k$ element-wise, as
\begin{equation}\label{eq:g_elementwise}
[g_k]_i = \frac{[f(z_k)]_i}{[z_k]_i}.
\end{equation}
Here $0/0$ should be interpreted as $0$. For example, the weight map $g_k$ associated with the ReLU function $f(\cdot)$, is a binary map which is a thresholded version of $z_k$,
\begin{equation}\label{eq:ActivationElementwise}
[g_k]_i =
\begin{cases}
1 & [z_k]_i > 0, \\
0 & [z_k]_i \leq 0.
\end{cases}
\end{equation}
This interpretation is visualized in Fig.~\ref{fig:scheme}(a) (bias not shown).


Since the nonlinear activations are what grants CNNs their ability to implement complex functions, here we propose to use \emph{learnable spatial activations}. That is, instead of the element-wise relation \eqref{eq:g_elementwise}, we propose to allow each element in $g_k$ to depend also on the spatial neighborhood of the corresponding element in $z_k$. Specifically, we introduce \emph{xUnit}, in which
\begin{equation}
[g_k]_i = \exp\{-[d_k]_i^2\},
\end{equation}
and
\begin{equation}
d_k = H_k \; \text{ReLU}(z_k),
\end{equation}
with $H_k$ denoting depth-wise convolution \cite{conv2ddepthwise}. The idea is to introduce (i)~nonlinearity (ReLU), (ii)~spatial processing (depth-wise convolution), and (iii)~construction of gating maps in the range $[0,1]$ (Gaussian). The depth-wise convolution applies a single filter to each input channel, and is significantly more efficient in terms of memory and computations than the multi-channel convolution popularly used in CNNs. Note that the filters $H_k$ have to be learned during training. To make the training stable, we also add batch-normalization layers~\cite{bn} before the ReLU and before the exponentiation. This is illustrated in Fig.~\ref{fig:scheme}(b).


Merely replacing ReLUs with xUnits clearly increases memory consumption and running times at test stage. This is mostly due to their convolutional operations (the exponent can be implemented using a look-up table). Specifically, an xUnit with a $d$-channel input and a $d$-channel output involving $r\times r$ filters, introduces an overhead of $(r^2+4)d$ parameters ($r\times r\times d$ for the depth-wise filters, and $2\times d$ for each batch-normalization layer). However, first, note that this overhead is relatively mild compared to the $r^2d^2$ parameters of each $r\times r\times d\times d$ convolutional layer. Second, in return to that overhead, xUnits provide a performance boost. This means that the same performance can be attained with less layers or with less channels per layer. Therefore, the important question is whether xUnits improve the overall \emph{tradeoff} between performance and number of parameters.

\begin{figure}
		\includegraphics[width=\columnwidth]{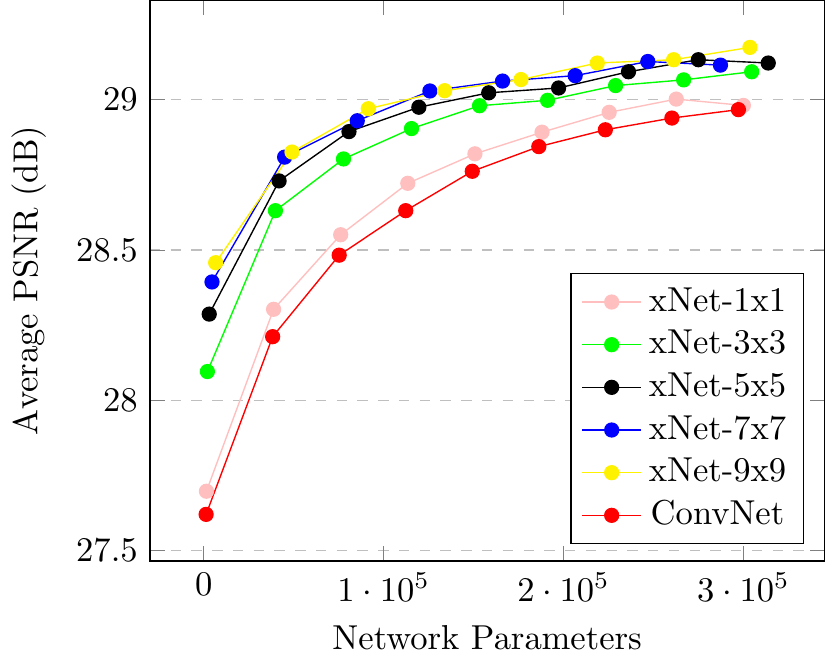}
	\caption{\textbf{Denoising performance vs.\@ number of parameters.} We compare a ConvNet composed of feed-forward Conv+BN+ReLU layers, with our xNet which comprises a sequence of Conv+xUnit layers. We gradually increase the number of layers for both nets and record the average PSNR obtained in denoising the BSD68 dataset with noise level $\sigma = 25$, as a function of the total number of parameters in the net. Training configurations are the same for both networks. Our xNet attains a much higher PSNR with the same number of parameters. Alternatively, it can achieve the same PSNR with roughly one third the number of ConvNet parameters.}\label{fig:params_vs_psnrs}
\end{figure}

\begin{figure}
\includegraphics[width=\columnwidth]{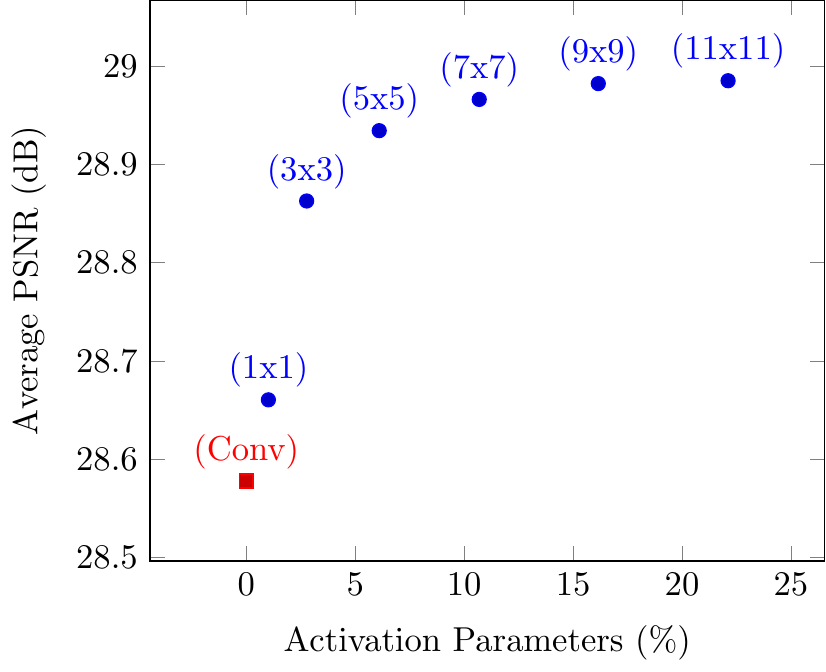}
\caption{\textbf{Denoising performance vs.\@ percentage of activation parameters.} Varying the supports of the xUnit filters improves performance, but also increases the overall number of parameters. Here, we show the average denoising PSNR as a function of the \emph{percentage of the overall parameters} invested in the xUnit activations, when the total number of parameters is constrained to $99,136$. This corresponds to a vertical cross section of the graph in Fig.~\ref{fig:params_vs_psnrs}. As can be seen, while conventional ConvNets invest $0\%$ of the parameters in the activations, this is clearly sub-optimal. A significant improvement in performance is obtained when investing about $20\%$ of the total parameters in the activations.}\label{fig:spatial_vs_psnrs}
\end{figure}

Figure~\ref{fig:params_vs_psnrs} shows the effect of using xUnits in a denoising task. Here, we trained two simple net architectures to remove additive Gaussian noise of standard deviation $\sigma=25$ from noisy images, using a varying number of layers. The first net is a traditional ConvNet architecture comprising a sequence of Conv+BN+ReLU layers. The second net, which we coin xNet, comprises a sequence of Conv+xUnit layers. In both nets, the regular convolutional layers (not the ones within the xUnits) comprise $64$ channel $3\times 3$ filters. For the xUnits, we varied the size of the depth-wise filters from $1\times1$ to $9\times 9$. We trained both nets on 400 images from the BSD dataset \cite{BSD} using residual learning (\ie learning to predict the noise and subtracting the noise estimate from the noisy image at test time). This has been shown to be advantageous for denoising in \cite{dncnn}. As can be seen in the figure, when the xUnit filters are $1\times 1$, the peak signal to noise ratio (PSNR) attained by xNet exceeds that of ConvNet by only a minor gap. In this case, the xUnits are not spatial. However, as the xUnits' filters become larger, xNet's performance begins to improve, for any given total number of parameters. Note, for example, that a 3 layer xNet with $9\times 9$ activations outperforms a 9 layer ConvNet, although having less than $1/3$ the number of parameters.  

\begin{figure*}
\begin{center}
\includegraphics[width=\textwidth]{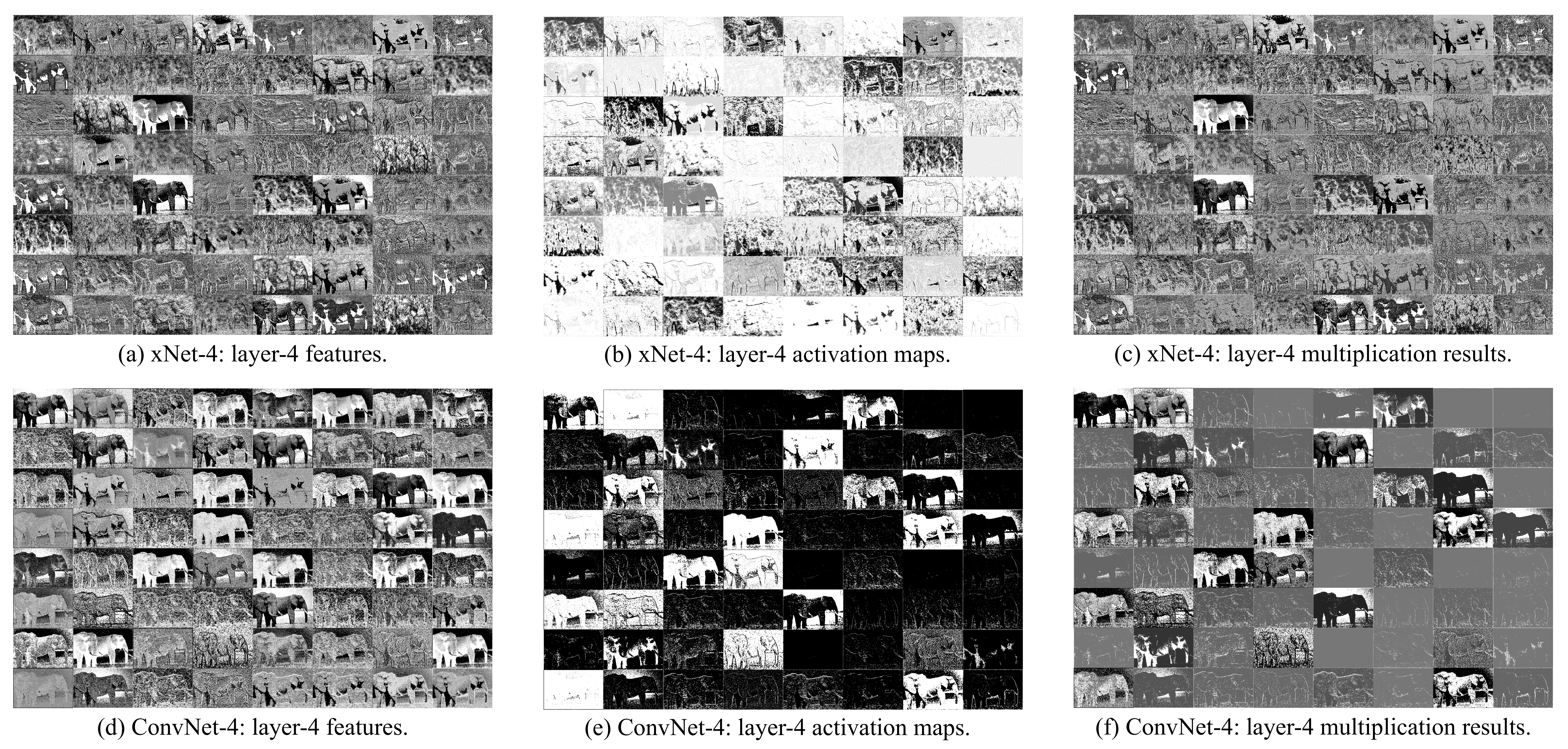}
\end{center}
\caption{\textbf{Visualization of xNet activations.} A 4-layer xNet and a 4-layer ConvNet (with ReLU activations) were trained to denoise images with noise level $\sigma$ = 25 using direct learning. The 64 feature maps, activation maps, and their products, at layer 4 are shown for both nets, when operating on the same input image. In xNet, each activation map is a spatial function of the corresponding feature map, whereas in ConvNet, it is a per-pixel function. As can be seen, the ConNet's activations are very sparse, while the xNet's activations are quite dense. Thus, it seems that in xNet, more feature maps participate in the denoising effort.}\label{fig:channels}
\end{figure*}

To further understand the performance-computation tradeoff when using spatial activations, Fig.~\ref{fig:spatial_vs_psnrs} shows a vertical cross section of the graph in Fig.~\ref{fig:params_vs_psnrs} at an overall of 99,136 parameters. Here, the PSNR is plotted against the percentage of parameters invested in the xUnit activations. In a traditional ConvNet, $0\%$ of the parameters are invested in the activations. As can be seen in the graph, this is clearly sub-optimal. In particular, the optimal percentage can be seen to be at least $22\%$, where the performance of xNet reaches a plateau. In fact, after around $15\%$ (corresponding to $9\times 9$ activation filters), the benefit in further increasing the filters' supports becomes relatively minor.


To gain intuition into the mechanism that allows xNet to achieve better results with less parameters, we depict in Fig.~\ref{fig:channels} the layer 4 feature maps $z_4$, weight (activation) maps $g_4$, and their products $x_5$, for a ConvNet and an xNet operating on the same noisy input image. Interestingly, we see that many more xNet activations are close to $1$ (white) than ConvNet activations. Thus, it seems that in xNet, more channels take part in the denoising effort. Moreover, it can be seen that the xNet weight maps are quite complex functions of the features, as opposed to the simple binarization function of the ReLUs in ConvNet.


\begin{figure}
\includegraphics[width=\columnwidth]{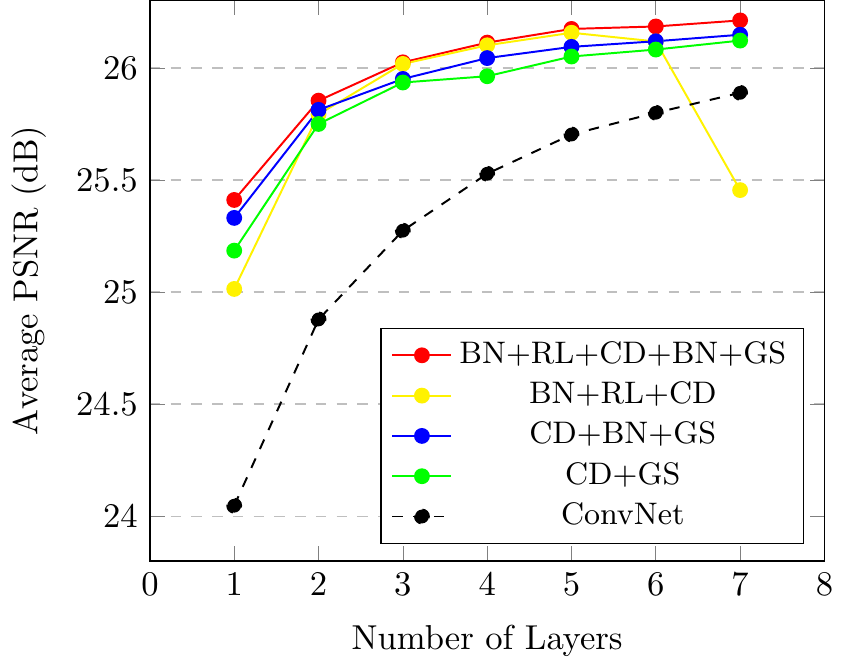}
\caption{\textbf{xUnit design comparison.} We compare various xUnit designs with a traditional ConvNet. We gradually increase the number of layers for all nets and record the average PSNR obtained in denoising the BSD68 dataset with noise level $\sigma=50$. Training configurations are the same for all nets. The suggested version, BN+RL+CD+BN+GS, attains the highest PSNR.}\label{fig:designs}
\end{figure}

Figure \ref{fig:designs} compares several alternative xUnit designs. The suggested design, which contains Batch Norm.~(BN), ReLU~(RL), Conv.~Depth-wise (CD) and Gaussian (GS), achieves the best results. However, note that all designs perform significantly better than a conventional ConvNet, indicating that the spatial processing (CD, which appears in all designs) contributes the most. Interestingly, the BN+RL+CD combination, which allows the weight maps to contain negative values, preforms quite similarly to our suggested design when the number of layers is small. Nonetheless, unlike the other designs, for a larger number of layers we experience gradients exploding during training. This highlights the importance of the Gaussian, which regulates training by keeping weights in the range $[0,1]$.


\section{Experiments and Applications}

\begin{table*}[t]
\begin{center}
\begin{tabular}{|c||c|c|c|c|c|c|}
\hline
Methods & BM3D & WNNM & EPLL & MLP & DnCNN-S & xDnCNN \\ \hline
\# of parameters & - & - & - & - & 555K & \textbf{303K} \\ \hline
$\sigma=25$ & 28.56 & 28.82 & 28.68 & 28.95 & \textbf{29.22} & 29.21 \\ \hline
$\sigma=50$ & 25.62 & 25.87 & 25.67 & 26.01 & 26.23 & \textbf{26.26} \\ \hline
\end{tabular}
\end{center}
\caption{\textbf{Denoising performance.} The average PSNR in [dB] attained by several state of the art denoising algorithms on the BSD68 dataset. As can be seen, our xDnCNN outperforms all non-CNN methods and achieves results that are on par with DnCNN, although having roughly $1/2$ the number of parameters.}\label{psnrs}
\end{table*}

\begin{figure*}
\begin{center}
\includegraphics[width=\textwidth]{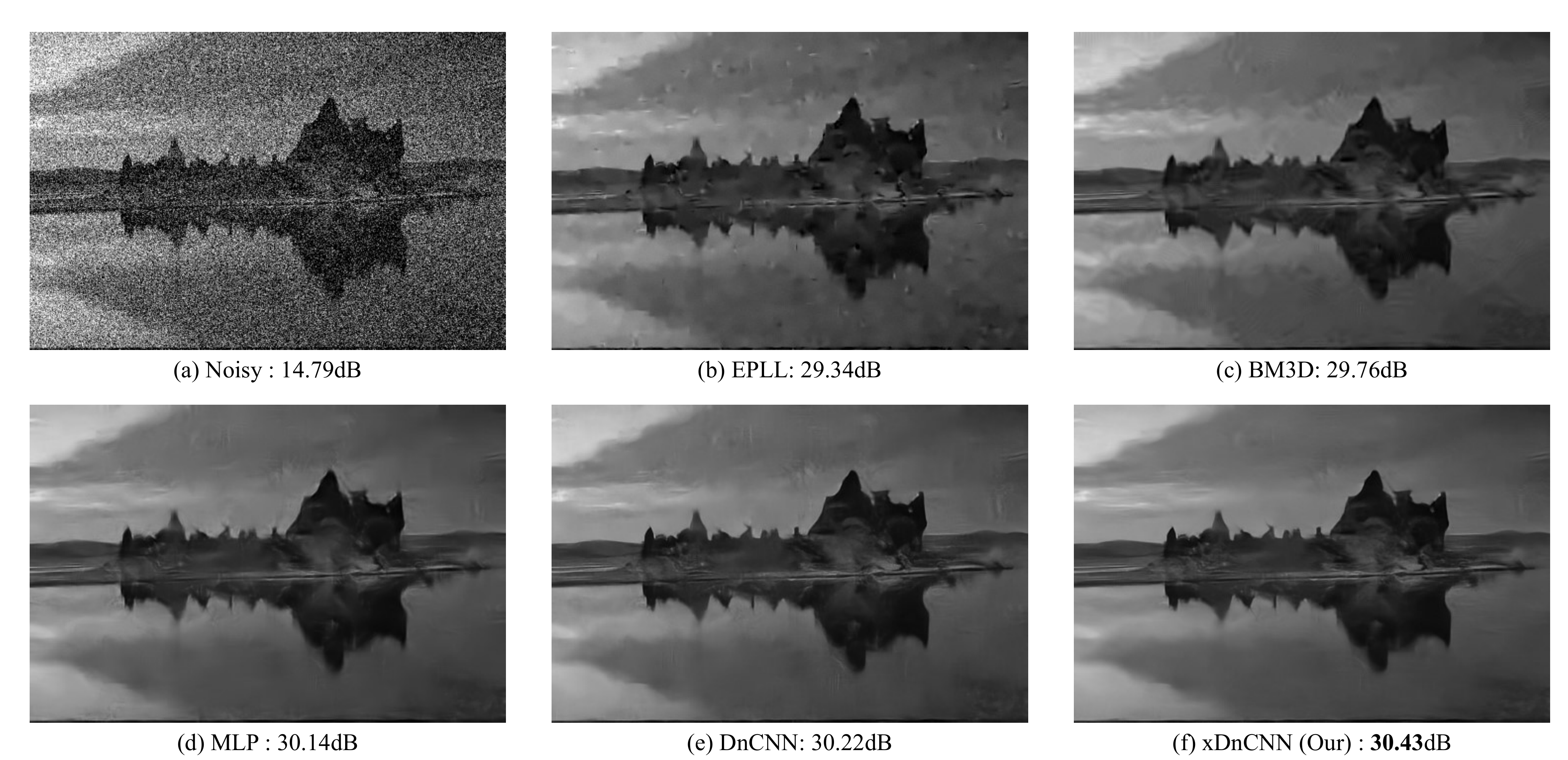}
\end{center}
\caption{ \textbf{Image denoising result.} Comparison of the denoised images produced by EPLL, BM3D, MLP, DnCNN and our xDnCNN for a noise level of $\sigma=50$. In contrast to the competing methods, our xDnCNN manages to restore more of the image details, and introduces no distracting artifacts. This is despite the fact that it has nearly half the number of parameters as DnCNN.}\label{denoising}
\end{figure*}

Our goal is to show that many small-scale and medium-scale state-of-the-art CNNs can be made almost $50\%$ smaller with xUnits, without incurring any degradation in performance.

We implemented the proposed architecture in Pytorch. We ran all experiments on a desktop computer with an Intel i5-6500 CPU and an Nvidia 1080Ti GPU. We used the Adam \cite{kingma2014adam} optimizer with its default settings for training the nets. We initialized the learning rate to $10^{-3}$ and gradually decreased it to $10^{-4}$ during training. We kept the mini-batch size fixed at 64. In all applications, we used $9\times 9$ depth-wise convolutions in the xUnits, and minimized the mean square error (MSE) over the training set. 


\subsection{Image Denoising}

\begin{figure*}
\begin{center}
\includegraphics[width=\textwidth]{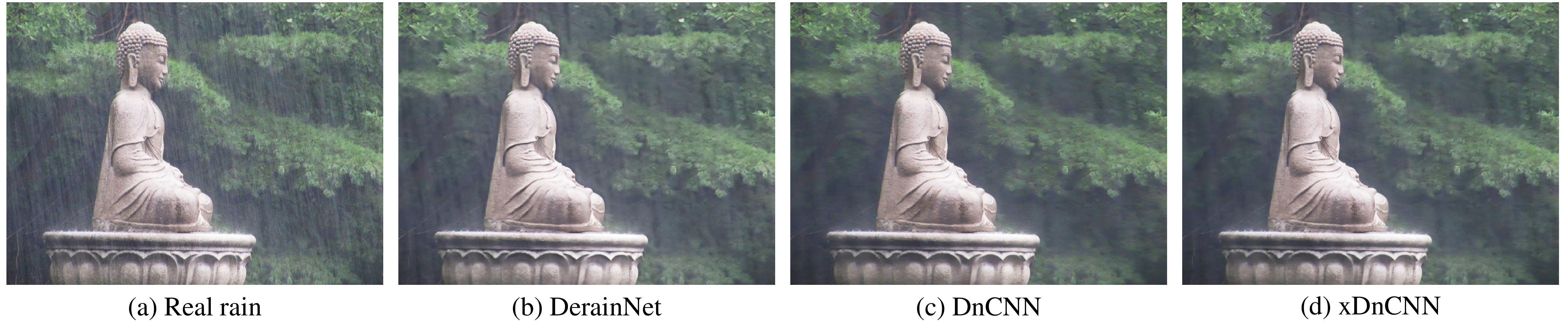}
\end{center}
\caption{\textbf{De-raining a real rain image.} Our xDnCNN deraining network manages to remove rain streaks from a real rainy image, significantly better than DerainNet. This is despite the fact that our net has only $40\%$ the number of parameters of DerainNet.}\label{fig:deraining}
\end{figure*}

We begin by illustrating the effectiveness of xUnits in image denoising. As a baseline architecture, we take the state-of-the-art DnCNN denoising network~\cite{dncnn}. We replace all ReLU layers with xUnit layers and reduce the number of convolutional layers from 17 to 9. We keep all convolutional layers with 64 channel $3\times 3$ filters, as in the original architecture. Our net, which we coin \emph{xDnCNN}, has only $54\%$ the number of parameters of DnCNN (303K for xDnCNN and 555K for DnCNN).

As in \cite{dncnn}, we train our net on 400 images. We use images from the Berkeley segmentation dataset (BSD) \cite{BSD}, enriched by random flipping and random cropping ($80\times80$). The noisy images are generated by adding Gaussian noise to the training images (different realization to each image). We examine the performance of our net at noise levels $\sigma=25,50$. Table~\ref{psnrs} compares the average PSNR attained by our xDnCNN to that attained by the original DnCNN (variant `S'), as well as to the state-of-the-art non-CNN denoising methods BM3D \cite{bm3d}, WNNM \cite{wnnm}, EPLL \cite{epll}, and MLP \cite{mlp}. The evaluation is performed on the BSD68 dataset \cite{bsd68}, a subset of 68 images from the BSD dataset, which is not included in the training set. As can be seen, our xDnCNN outperforms all non-CNN denoising methods and achieves results that are on par with DnCNN. This is despite the fact that xDnCNN is nearly half the size of DnCNN in terms of number of parameters. The superiority of our method becomes more significant as the noise level increases. At a noise level of $\sigma = 50$, our method achieves the highest PSNR values on $57$ out of the $68$ images in the dataset.

Figure~\ref{denoising} shows an example denoising result obtained with xDnCNN, compared with BM3D, EPLL, MLP and DnCNN-s, for a noise level of $\sigma = 50$. As can be seen, our xDnCNN best reconstructs the fine details and barely introduces any distracting artifacts. In contrast, all the other methods (including DnCNN), introduce unpleasing distortions.

\subsection{Single image rain removal}

Next, we use the same architecture in the task of removing rain streaks from a single image. We only introduce one modification to our denoising xDnCNN, which is to work on three channel (RGB) input images and to output three channel images. This results in a network with 306K  parameters. We compare our results to a 3-channel input 3-channel output DnCNN version and to DerainNet \cite{derainnet}, a network with $753$K parameters, which comprises three convolutional layers: $16\times 16\times 512$, $1\times 1\times 512$ and $8\times 8\times 3$, respectively. Similarly to denoising, we learn the residual mapping between a rainy image and a clean image. Training is performed on the dataset of DerainNet \cite{derainnet}, which contains 4900 pairs of clean and synthetically generated rainy images. However, we evaluate our net on the Rain12 dataset \cite{li2016rain}, which contains 12 artificially generated images. Although the training data is quite different from the test data, our xDnCNN performs significantly better than DnCNN and DerainNet, as shown in Table \ref{deraining_psnr}. This behavior is also seen when de-raining real images. As can be seen in Fig.~\ref{fig:deraining}, xDnCNN perform significantly better in cleaning actual rain streaks. We thus conclude that xDnCNN is far more robust to different rain appearances, while maintaining its efficiency. Pay attention that our xDnCNN deraining net has only $40\%$ the number of parameters of DerainNet and only $55\%$ the number of parameters of DnCNN.

\begin{table}[t]
\begin{center}
\begin{tabular}{|c||c|c|c|}
\hline
Methods & De-rainNet & DnCNN & xDnCNN \\ \hline
\# of parameters & 753K & 558K &\textbf{306K} \\ \hline
PSNR [dB] & 28.94 & 30.90 & \textbf{31.17} \\ \hline
\end{tabular}
\end{center}
\caption{\textbf{De-raining performance on the Rain12 dataset.} Our xDnCNN attains a significantly higher PSNR than DnCNN and De-rainNet, with significantly less parameters.}\label{deraining_psnr}
\end{table}

\begin{table*}[t]
\begin{center}
\begin{tabular}{|c||c|c|c||c|c|}
\hline
Methods & SRCNN & xSRCNNc & xSRCNNf & SRResNet & xSRResNet \\ \hline
\# of parameters & 57K & 44K & \textbf{32K} & 1.546M & \textbf{1.155M} \\ \hline
$3\times$ & 28.41 & \textbf{28.54} & 28.53 & - & - \\ \hline
$4\times$ & 26.90 & 27.04 & \textbf{27.06} & 27.58 & \textbf{27.61} \\ \hline
\end{tabular}
\end{center}
\caption{\textbf{Super-resolution performance.} The average PSNR in [dB] attained in the task of $3\times$ and $4\times$ SR on BSD100 dataset. SRCNN was trained on $4\times10^5$ training examples, whereas our xSRCNN models were trained on only $491$ images. SRResNet was trained on $3.5\times10^5$ training examples, whereas our xSRResNet was trained on $2.5\times10^4$ examples.}
\label{sr}
\end{table*}

\begin{figure*}[h]
\centering
\includegraphics[width=1\textwidth]{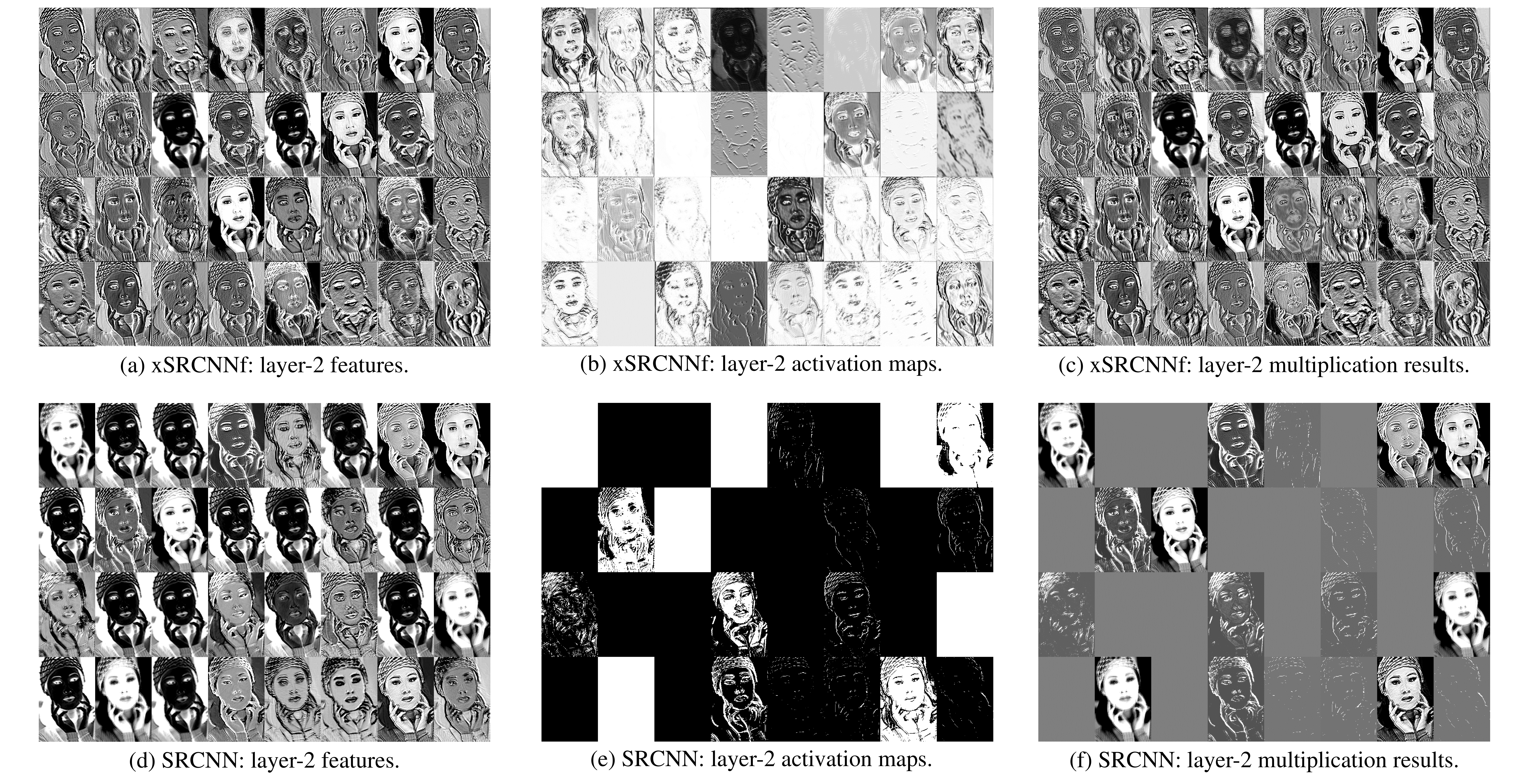}
\caption{\textbf{Visualization of SRCNN and xSRCNNf activations.} The 32 features maps, activation maps, and their products at layer 2 are shown for both nets, when operating on the same input image for magnification $3\times$. As can be seen, the activations of SRCNN are sparse, while those of xSRCNNf are dense. Thus, more feature maps participate in the reconstruction. This provides an explanation for the superiority of xSRCNNf over SRCNN in terns of PSNR.}\label{fig:SRchannels}
\end{figure*}

\subsection{Single image super resolution}
Our xUnit activations can be also applied in single image super resolution. We illustrate this with the state-of-the-art SRResNet architecture \cite{srgan} and with the very small SRCNN \cite{srcnn} model. For SRResNet, we replace the PReLU activations in the residual blocks by xUnits, and reduce the number of residual blocks from 16 to 10. This variant, which we coin xSRResNet, has only 75\% the number of parameters of SRResNet (1.546M for SRResNet and 1.155M for xSRResNet). The SRCNN architecture contains three convolutional layers: $9\times 9\times 64$, $5\times 5\times 32$ and $5\times 5\times 1$. Here, we study two different modifications to SRCNN, where we replace the two ReLU layers with xUnit layers. In the first modification, we reduce the size of the filters in the middle layer from $5\times 5\times 32$ to $3\times 3\times 32$. This variant, which we coin xSRCNNf, has only $56\%$ the number of parameters of SRCNN ($32$K for xSRCNNf and $57$K for SRCNN). In the second modification, we reduce the number of channels in the second layer from $64$ to $42$. This variant, which we coin xSRCNNc, has only $77\%$ the number of parameters of SRCNN (44K for xSRCNNc and 57K for SRCNN).

The original SRCNN and SRResNet models were trained on about $400,000$ images from ImageNet~\cite{russakovsky2015imagenet}. Here, we train our models on much smaller datasets. Specifically, for xSRCNN we use only $91$ images from \cite{91dataset} and $400$ images from BSD. For xSRResNet, we use $25,000$ images from the Mirflickr25k \cite{flicker} dataset. We augment the data by random flipping and random cropping.

Table~\ref{sr} reports the results attained by all the models on BSD100 dataset. As can be seen, our models attain results that are on par with the original SRCNN and SRResNet, although being much smaller and trained on a significantly smaller number of images. Note that our xSRCNNf has less parameters than xSRCNNc. This may suggests that a better way to discard parameters in xNets is by reducing filter sizes, rather than reducing channels. A possible explanation is that the $9\times 9$ filters within the xUnits can partially compensate for the small support of the filters in the convolutional layers. However, the fact that discarding channels can also provide a significant reduction in parameters at the same performance, indicates that the channels in an xNet are more effective than those in ConvNets with per-pixel activations.

Figure~\ref{fig:SRchannels} shows the layer 2 feature maps, activation maps, and their products for both SRCNN and our xSRCNNf. As in the case of denoising, we can see that in xSRCNN, many more feature maps participate in the reconstruction effort compared to SRCNN. This provides a possible explanation to its ability to perform well with smaller filters (or with less channels).

\section{Conclusion}
Popular CNN architectures use simple nonlinear activation units (\eg ReLUs), which operate pixel-wise on the feature maps. In this paper, we demonstrated that CNNs can greatly benefit from incorporating learnable spatial connections within the activation units. While these spatial connections introduce additional parameters to the net, they significantly improve its performance. Overall, the tradeoff between performance and number of parameters, is substantially improved. We illustrated how our approach can reduce the size of several state-of-the-art CNN models for denoising, de-raining and super-resolution, which are already considered to be very small, by almost 50\%. This is  without incurring any degradation in performance. 

\noindent\textbf{Acknowledgements\ \ \ }
This research was supported in part by an Alon Fellowship and by the Ollendorf Foundation.

{\small
\bibliography{egbib}

\begin{thebibliography}{10}\itemsep=-1pt

\bibitem{segnet}
V.~Badrinarayanan, A.~Kendall, and R.~Cipolla.
\newblock Segnet: A deep convolutional encoder-decoder architecture for image
  segmentation.
\newblock {\em arXiv preprint arXiv:1511.00561}, 2015.

\bibitem{bae2016beyond}
W.~Bae, J.~Yoo, and J.~C. Ye.
\newblock Beyond deep residual learning for image restoration: Persistent
  homology-guided manifold simplification.
\newblock {\em arXiv preprint arXiv:1611.06345}, 2016.

\bibitem{mlp}
H.~C. Burger, C.~J. Schuler, and S.~Harmeling.
\newblock Image denoising: Can plain neural networks compete with bm3d?
\newblock In {\em Computer Vision and Pattern Recognition (CVPR), 2012 IEEE
  Conference on}, pages 2392--2399. IEEE, 2012.

\bibitem{cai2016dehazenet}
B.~Cai, X.~Xu, K.~Jia, C.~Qing, and D.~Tao.
\newblock Dehazenet: An end-to-end system for single image haze removal.
\newblock {\em IEEE Transactions on Image Processing}, 25(11):5187--5198, 2016.

\bibitem{elu}
D.-A. Clevert, T.~Unterthiner, and S.~Hochreiter.
\newblock Fast and accurate deep network learning by exponential linear units
  (elus).
\newblock {\em arXiv preprint arXiv:1511.07289}, 2015.

\bibitem{courbariaux2016binarized}
M.~Courbariaux, I.~Hubara, D.~Soudry, R.~El-Yaniv, and Y.~Bengio.
\newblock Binarized neural networks: Training deep neural networks with weights
  and activations constrained to+ 1 or-1.
\newblock {\em arXiv preprint arXiv:1602.02830}, 2016.

\bibitem{bm3d}
K.~Dabov, A.~Foi, V.~Katkovnik, and K.~Egiazarian.
\newblock Image denoising by sparse 3-d transform-domain collaborative
  filtering.
\newblock {\em IEEE Transactions on image processing}, 16(8):2080--2095, 2007.

\bibitem{dong2014learning}
C.~Dong, C.~C. Loy, K.~He, and X.~Tang.
\newblock Learning a deep convolutional network for image super-resolution.
\newblock In {\em European Conference on Computer Vision}, pages 184--199.
  Springer, 2014.

\bibitem{srcnn}
C.~Dong, C.~C. Loy, K.~He, and X.~Tang.
\newblock Image super-resolution using deep convolutional networks.
\newblock {\em IEEE transactions on pattern analysis and machine intelligence},
  38(2):295--307, 2016.

\bibitem{derainnet}
X.~Fu, J.~Huang, X.~Ding, Y.~Liao, and J.~Paisley.
\newblock Clearing the skies: A deep network architecture for single-image rain
  removal.
\newblock {\em IEEE Transactions on Image Processing}, 26(6):2944--2956, 2017.

\bibitem{relu}
X.~Glorot, A.~Bordes, and Y.~Bengio.
\newblock Deep sparse rectifier neural networks.
\newblock In {\em Proceedings of the Fourteenth International Conference on
  Artificial Intelligence and Statistics}, pages 315--323, 2011.

\bibitem{wnnm}
S.~Gu, L.~Zhang, W.~Zuo, and X.~Feng.
\newblock Weighted nuclear norm minimization with application to image
  denoising.
\newblock In {\em Proceedings of the IEEE Conference on Computer Vision and
  Pattern Recognition}, pages 2862--2869, 2014.

\bibitem{he2015delving}
K.~He, X.~Zhang, S.~Ren, and J.~Sun.
\newblock Delving deep into rectifiers: Surpassing human-level performance on
  imagenet classification.
\newblock In {\em Proceedings of the IEEE international conference on computer
  vision}, pages 1026--1034, 2015.

\bibitem{resnet}
K.~He, X.~Zhang, S.~Ren, and J.~Sun.
\newblock Deep residual learning for image recognition.
\newblock In {\em The IEEE Conference on Computer Vision and Pattern
  Recognition (CVPR)}, June 2016.

\bibitem{he2016identity}
K.~He, X.~Zhang, S.~Ren, and J.~Sun.
\newblock Identity mappings in deep residual networks.
\newblock In {\em European Conference on Computer Vision}, pages 630--645.
  Springer, 2016.

\bibitem{conv2ddepthwise}
A.~G. Howard, M.~Zhu, B.~Chen, D.~Kalenichenko, W.~Wang, T.~Weyand,
  M.~Andreetto, and H.~Adam.
\newblock Mobilenets: Efficient convolutional neural networks for mobile vision
  applications.
\newblock {\em arXiv preprint arXiv:1704.04861}, 2017.

\bibitem{densenet}
G.~Huang, Z.~Liu, K.~Q. Weinberger, and L.~van~der Maaten.
\newblock Densely connected convolutional networks.
\newblock {\em arXiv preprint arXiv:1608.06993}, 2016.

\bibitem{flicker}
M.~J. Huiskes and M.~S. Lew.
\newblock The mir flickr retrieval evaluation.
\newblock In {\em Proceedings of the 1st ACM international conference on
  Multimedia information retrieval}, pages 39--43. ACM, 2008.

\bibitem{bn}
S.~Ioffe and C.~Szegedy.
\newblock Batch normalization: Accelerating deep network training by reducing
  internal covariate shift.
\newblock In {\em International Conference on Machine Learning}, pages
  448--456, 2015.

\bibitem{formresnet}
J.~Jiao, W.-C. Tu, S.~He, and R.~W. Lau.
\newblock Formresnet: Formatted residual learning for image restoration.
\newblock In {\em Computer Vision and Pattern Recognition Workshops (CVPRW),
  2017 IEEE Conference on}, pages 1034--1042. IEEE, 2017.

\bibitem{kim2016accurate}
J.~Kim, J.~Kwon~Lee, and K.~Mu~Lee.
\newblock Accurate image super-resolution using very deep convolutional
  networks.
\newblock In {\em Proceedings of the IEEE Conference on Computer Vision and
  Pattern Recognition}, pages 1646--1654, 2016.

\bibitem{kingma2014adam}
D.~Kingma and J.~Ba.
\newblock Adam: A method for stochastic optimization.
\newblock {\em arXiv preprint arXiv:1412.6980}, 2014.

\bibitem{srgan}
C.~Ledig, L.~Theis, F.~Husz{\'a}r, J.~Caballero, A.~Cunningham, A.~Acosta,
  A.~Aitken, A.~Tejani, J.~Totz, Z.~Wang, et~al.
\newblock Photo-realistic single image super-resolution using a generative
  adversarial network.
\newblock {\em arXiv preprint arXiv:1609.04802}, 2016.

\bibitem{lefkimmiatis2016non}
S.~Lefkimmiatis.
\newblock Non-local color image denoising with convolutional neural networks.
\newblock {\em arXiv preprint arXiv:1611.06757}, 2016.

\bibitem{li2017all}
B.~Li, X.~Peng, Z.~Wang, J.~Xu, and D.~Feng.
\newblock An all-in-one network for dehazing and beyond.
\newblock {\em arXiv preprint arXiv:1707.06543}, 2017.

\bibitem{li2016rain}
Y.~Li, R.~T. Tan, X.~Guo, J.~Lu, and M.~S. Brown.
\newblock Rain streak removal using layer priors.
\newblock In {\em Proceedings of the IEEE Conference on Computer Vision and
  Pattern Recognition}, pages 2736--2744, 2016.

\bibitem{lim2017enhanced}
B.~Lim, S.~Son, H.~Kim, S.~Nah, and K.~M. Lee.
\newblock Enhanced deep residual networks for single image super-resolution.
\newblock In {\em The IEEE Conference on Computer Vision and Pattern
  Recognition (CVPR) Workshops}, 2017.

\bibitem{maas2013rectifier}
A.~L. Maas, A.~Y. Hannun, and A.~Y. Ng.
\newblock Rectifier nonlinearities improve neural network acoustic models.
\newblock In {\em Proc. ICML}, volume~30, 2013.

\bibitem{BSD}
D.~Martin, C.~Fowlkes, D.~Tal, and J.~Malik.
\newblock A database of human segmented natural images and its application to
  evaluating segmentation algorithms and measuring ecological statistics.
\newblock In {\em Proc. 8th Int'l Conf. Computer Vision}, volume~2, pages
  416--423, July 2001.

\bibitem{noroozi2017motion}
M.~Noroozi, P.~Chandramouli, and P.~Favaro.
\newblock Motion deblurring in the wild.
\newblock {\em arXiv preprint arXiv:1701.01486}, 2017.

\bibitem{orr2003neural}
G.~B. Orr and K.-R. M{\"u}ller.
\newblock {\em Neural networks: tricks of the trade}.
\newblock Springer, 2003.

\bibitem{parkhi2015deep}
O.~M. Parkhi, A.~Vedaldi, A.~Zisserman, et~al.
\newblock Deep face recognition.
\newblock In {\em BMVC}, volume~1, page~6, 2015.

\bibitem{portilla2003image}
J.~Portilla, V.~Strela, M.~J. Wainwright, and E.~P. Simoncelli.
\newblock Image denoising using scale mixtures of gaussians in the wavelet
  domain.
\newblock {\em IEEE Transactions on Image processing}, 12(11):1338--1351, 2003.

\bibitem{remez2017deep1}
T.~Remez, O.~Litany, R.~Giryes, and A.~M. Bronstein.
\newblock Deep class aware denoising.
\newblock {\em arXiv preprint arXiv:1701.01698}, 2017.

\bibitem{remez2017deep}
T.~Remez, O.~Litany, R.~Giryes, and A.~M. Bronstein.
\newblock Deep class-aware image denoising.
\newblock In {\em Sampling Theory and Applications (SampTA), 2017 International
  Conference on}, pages 138--142. IEEE, 2017.

\bibitem{remez2017deep2}
T.~Remez, O.~Litany, R.~Giryes, and A.~M. Bronstein.
\newblock Deep convolutional denoising of low-light images.
\newblock {\em arXiv preprint arXiv:1701.01687}, 2017.

\bibitem{ren2016single}
W.~Ren, S.~Liu, H.~Zhang, J.~Pan, X.~Cao, and M.-H. Yang.
\newblock Single image dehazing via multi-scale convolutional neural networks.
\newblock In {\em European Conference on Computer Vision}, pages 154--169.
  Springer, 2016.

\bibitem{unet}
O.~Ronneberger, P.~Fischer, and T.~Brox.
\newblock U-net: Convolutional networks for biomedical image segmentation.
\newblock In {\em International Conference on Medical Image Computing and
  Computer-Assisted Intervention}, pages 234--241. Springer, 2015.

\bibitem{foe}
S.~Roth and M.~J. Black.
\newblock Fields of experts: A framework for learning image priors.
\newblock In {\em Computer Vision and Pattern Recognition, 2005. CVPR 2005.
  IEEE Computer Society Conference on}, volume~2, pages 860--867. IEEE, 2005.

\bibitem{bsd68}
S.~Roth and M.~J. Black.
\newblock Fields of experts.
\newblock {\em International Journal of Computer Vision}, 82(2):205--229, 2009.

\bibitem{rudin1992nonlinear}
L.~I. Rudin, S.~Osher, and E.~Fatemi.
\newblock Nonlinear total variation based noise removal algorithms.
\newblock {\em Physica D: Nonlinear Phenomena}, 60(1-4):259--268, 1992.

\bibitem{russakovsky2015imagenet}
O.~Russakovsky, J.~Deng, H.~Su, J.~Krause, S.~Satheesh, S.~Ma, Z.~Huang,
  A.~Karpathy, A.~Khosla, M.~Bernstein, et~al.
\newblock Imagenet large scale visual recognition challenge.
\newblock {\em International Journal of Computer Vision}, 115(3):211--252,
  2015.

\bibitem{facenet}
F.~Schroff, D.~Kalenichenko, and J.~Philbin.
\newblock Facenet: A unified embedding for face recognition and clustering.
\newblock In {\em Proceedings of the IEEE Conference on Computer Vision and
  Pattern Recognition}, pages 815--823, 2015.

\bibitem{shi2016real}
W.~Shi, J.~Caballero, F.~Husz{\'a}r, J.~Totz, A.~P. Aitken, R.~Bishop,
  D.~Rueckert, and Z.~Wang.
\newblock Real-time single image and video super-resolution using an efficient
  sub-pixel convolutional neural network.
\newblock In {\em Proceedings of the IEEE Conference on Computer Vision and
  Pattern Recognition}, pages 1874--1883, 2016.

\bibitem{vgg}
K.~Simonyan and A.~Zisserman.
\newblock Very deep convolutional networks for large-scale image recognition.
\newblock {\em arXiv preprint arXiv:1409.1556}, 2014.

\bibitem{timofte2017ntire}
R.~Timofte, E.~Agustsson, L.~Van~Gool, M.-H. Yang, L.~Zhang, B.~Lim, S.~Son,
  H.~Kim, S.~Nah, K.~M. Lee, et~al.
\newblock Ntire 2017 challenge on single image super-resolution: Methods and
  results.
\newblock In {\em Computer Vision and Pattern Recognition Workshops (CVPRW),
  2017 IEEE Conference on}, pages 1110--1121. IEEE, 2017.

\bibitem{91dataset}
J.~Yang, J.~Wright, T.~S. Huang, and Y.~Ma.
\newblock Image super-resolution via sparse representation.
\newblock {\em IEEE transactions on image processing}, 19(11):2861--2873, 2010.

\bibitem{dncnn}
K.~Zhang, W.~Zuo, Y.~Chen, D.~Meng, and L.~Zhang.
\newblock Beyond a gaussian denoiser: Residual learning of deep cnn for image
  denoising.
\newblock {\em IEEE Transactions on Image Processing}, 2017.

\bibitem{zhang2017learning}
K.~Zhang, W.~Zuo, S.~Gu, and L.~Zhang.
\newblock Learning deep cnn denoiser prior for image restoration.
\newblock {\em arXiv preprint arXiv:1704.03264}, 2017.

\bibitem{epll}
D.~Zoran and Y.~Weiss.
\newblock From learning models of natural image patches to whole image
  restoration.
\newblock In {\em Computer Vision (ICCV), 2011 IEEE International Conference
  on}, pages 479--486. IEEE, 2011.

\end{thebibliography}
\bibliographystyle{ieee}
}

\end{document}